\documentclass[runningheads]{llncs}

\usepackage[colorlinks=true, urlcolor=blue, citecolor=blue, filecolor=blue, anchorcolor=blue, linkcolor=blue]{hyperref}
\usepackage{stackrel}
\usepackage{multirow}
\usepackage{makecell}
\usepackage{siunitx}
\usepackage{booktabs}

\usepackage{graphicx}
\usepackage{url}

\begin{document}

\title{RuMedBench: A Russian Medical Language Understanding Benchmark}


\author{
Pavel Blinov\inst{} \and
Arina Reshetnikova\inst{} \and Aleksandr Nesterov\inst{} \and Galina Zubkova\inst{} \and
Vladimir Kokh\inst{}
}

\authorrunning{P. Blinov et al.}

\institute{Sber Artificial Intelligence Laboratory, Moscow, Russia \\
\email{\{Blinov.P.D, AAnReshetnikova, AINesterov, GVZubkova, Kokh.V.N\}@sberbank.ru} \\
}

\maketitle 

\begin{abstract}
The paper describes the open Russian medical language understanding benchmark covering several task types (classification, question answering, natural language inference, named entity recognition) on a number of novel text sets. Given the sensitive nature of the data in healthcare, such a benchmark partially closes the problem of Russian medical dataset absence. We prepare the unified format labeling, data split, and evaluation metrics for new tasks. The remaining tasks are from existing datasets with a few modifications. A single-number metric expresses a model's ability to cope with the benchmark. Moreover, we implement several baseline models, from simple ones to neural networks with transformer architecture, and release the code. Expectedly, the more advanced models yield better performance, but even a simple model is enough for a decent result in some tasks. Furthermore, for all tasks, we provide a human evaluation. Interestingly the models outperform humans in the large-scale classification tasks. However, the advantage of natural intelligence remains in the tasks requiring more knowledge and reasoning.

\keywords{Natural Language Processing \and Benchmark \and Russian Medical Data \and EHR \and BERT.}
\end{abstract}

\section{Introduction}
In recent years deep neural network models have shown their effectiveness in solving multiple general Natural Language Processing (NLP) tasks. The progress in language model development actively extends to more narrow specific domains like medicine and healthcare. This trend intensified by the ongoing process of health industry digital transformation producing large volumes of data and posing challenging tasks requiring intelligent methods for processing. The primary source of text data in the medical domain is Electronic Health Records (EHRs). A series of models with novel BERT~\cite{devlin2018bert} architecture, e.g., ClinicalBERT~\cite{alsentzer-etal-2019-publicly}, BioBERT~\cite{lee2020biobert}, BlueBERT~\cite{peng2019transfer}, successfully applied to a range of healthcare and biomedical domain tasks.

In Artificial Intelligence (AI) research field, similar tasks often are grouped to a special benchmark containing a set of formalized Machine Learning (ML) problems with defined input data and performance metrics. For example, ImageNet~\cite{deng2009imagenet} benchmark for image classification or General Language Understanding Evaluation (GLUE) benchmark~\cite{wang2018glue}. The comparison of human and ML-model performances allows measuring the progress in a particular field. Thus, a domain benchmark is vital for AI research standardization and prerequisites for model development and evaluation.

Language dependency is a serious obstacle for NLP models development. An overwhelming majority of NLP research conducts on English datasets. It induces inequality in resources and benchmarks available for other languages as Russian. Although recently the general benchmark for Russian language understanding evaluation RussianSuperGLUE~\cite{russiansuperglue} was proposed, there is still no similar test for the medical domain. This work tries to close such a problem and propose the Russian medical language understanding benchmark. Two issues complicate the creation of such a benchmark. The first one is the private nature of textual medical data. For example, EHRs contain notes about patient identity, disease and treatment details, etc. Therefore, it is hard to automate the anonymization process, and it requires a thorough manual look-through. The second issue is the lack of annotations for a target task, which often involves the tedious hand labeling of samples.

In this paper, we make the following contributions:
\begin{itemize}
\item Introduce \emph{RuMedBench}, the first comprehensive open \emph{Ru}ssian \emph{Med}ical \linebreak language understanding \emph{Bench}mark. The RuMedBench contains five tasks based on four medical text types. The overall score allows to rank models on their ability to cope with medical specific tasks;
\item Propose three new datasets and tasks: \emph{RuMedTop3} for diagnosis prediction, \emph{RuMedSymptomRec} for symptoms recommendation, \emph{RuMedDaNet} for medical Question Answering (QA). Present \emph{RuMedNLI}, the translated version of the medical Natural Language Inference (NLI) task. Include Named Entity Recognition (NER) task on an existing dataset for broader ML task types coverage;
\item Perform a thorough comparison of several baselines from essential NLP approaches and estimate human performance for all proposed tasks.
\end{itemize}

The \emph{RuMedBench} data and baseline models source code available in this repository: \url{https://github.com/pavel-blinov/RuMedBench}.

\section{Related Work} \label{relat_sect}
The most known representatives of NLP benchmarks are GLUE~\cite{wang2018glue} and SuperGLUE~\cite{wang2019superglue}. The latter is the successor of the former, proposed with more challenging tasks to keep up with pacing progress in the NLP area. Both of them are designed for English general language understanding evaluation. For the Russian language, the analog of this benchmark is RussianSuperGLUE~\cite{russiansuperglue}. All those benchmarks include only single- or pair-centric text tasks. In this work, we also mostly use such task types but include a tokens-level problem. Because a typical medical text is full of specific terms, we want the benchmark to check models' ability to extract such terms and entities.

As already mentioned, medical privacy is one of the main difficulties for new datasets and relevant tasks creation. The publication in 2016 of the Medical Information Mart for Intensive Care (MIMIC-III) database~\cite{johnson2016mimic} led to active researches and a plethora of tasks appearance. In 2017, Purushotham et al.~\cite{purushotham2017benchmark} proposed a benchmark of three tasks: forecasting length of stay, mortality, and ICD-9 code group prediction. Only the last task overlaps with our ones as we focus on text data. The authors use 20 diagnosis groups to classify an intensive care unit admission. In our similar task, we use the newer 10th version of the International Classification of Diseases (ICD)~\cite{icd10}, the target set of predicted codes is five times larger, and input texts are from outpatient notes.

Next, Peng et al.~\cite{peng2019transfer} introduced the Biomedical Language Understanding Evaluation (BLUE) benchmark on ten datasets. Also, substantial work has been done in~\cite{lewis2020pretrained}, where the authors present a large-scale study across 18 biomedical and clinical NLP tasks. Our work is similar to these papers by the scope of task types. Although our test is inferior in the number of used datasets and tasks but differs positively by the broader range of medical text types. Moreover, our work has one intersection with \cite{peng2019transfer} and \cite{lewis2020pretrained}, a natural language inference task on the MedNLI~\cite{romanov2018lessons} dataset. For this, we translated original data (see Section~\ref{nli_sect}). That allows to test models in a similar task and also produced counterpart medical corpus.

All the above-discussed works refer to tasks in English. Recently similar works appeared for other languages, for example, a Chinese Biomedical Language Understanding Evaluation (CBLUE) benchmark~\cite{zhang2021cblue}. However, that cannot be said for Russian, where only a few separate datasets exist. We can note an interesting work of Tutubalina et al.~\cite{10.1093/bioinformatics/btaa675} on the NER task. Due to data availability, we incorporate this task in our benchmark. Another paper~\cite{shelmanov2015information} on information extraction from EHRs contains only 112 fully annotated texts. The data and labels are under user agreement license. In comparison to~\cite{shelmanov2015information}, EHR-related data in our work are open and by one order of magnitude larger in size (see Section~\ref{top3_sect}). Therefore our work is the first to introduce several new open Russian medical datasets and tasks: two for large-scale classification on outpatient notes, one for question answering, and one for natural language inference.

\section{Tasks and Data}
Medicine probably is one of the oldest areas of human knowledge containing a broad range of disciplines. It is a priori impossible to cover whole aspects of this area in a single benchmark. For this work, we omit other than text modalities of medical tasks.

All datasets in this study are prepared in compliance with fundamental principles of privacy and ethics. Table~\ref{tasks_data} lists the \emph{RuMedBench} sub-tasks along with some data characteristics and metrics. \emph{Accuracy} is the mandatory metric for all tasks, with additional metrics in exceptional cases. To estimate models' abilities in Russian medical language understanding, we propose to infer an overall score as mean over task metric values (with prior averaging in the case of two metrics).

\begin{table}
\centering
\caption{The \emph{RuMedBench} tasks.} \label{tasks_data}
\begin{tabular}{l l l c l l l c l}
\hline 
& Name & Type & Metrics & Train & Dev & Test & \begin{tabular}{@{}c@{}}\# of tokens / sents.\\Avg tokens per input \end{tabular} & Text type \\
\hline
\multirow{8}{*}{\rotatebox[origin=c]{90}{RuMed*}} & \multicolumn{1}{l}{Top3} & \multirow{3}{*}{Classif.} & \multirow{3}{*}{\begin{tabular}{c}Acc \\ Hit@3\end{tabular}} & 4,690 & 848 & 822 &
\begin{tabular}{@{}c@{}} 164,244 / 13,701 \\ 25 \end{tabular} 
& \multirow{3}{*}{\begin{tabular}{@{}l@{}}Outpatient \\ notes\end{tabular}} \\
\cline{2-2} \cline{5-8}
& \multicolumn{1}{l}{SymptomRec} & & & 2,470 & 415 & 415 & \begin{tabular}{@{}c@{}} 89,944 / 7,036 \\ 27 \end{tabular} & \\
\cline{2-9}
& \multicolumn{1}{l}{DaNet} & QA & \multirow{3}{*}{Acc} & 1,052 & 256 & 256 & \begin{tabular}{@{}c@{}} 102,708 / 7,307 \\ 65 \end{tabular} & \begin{tabular}{@{}l@{}}Misc. \\ medical\end{tabular} \\
\cline{2-3} \cline{5-9}
& \multicolumn{1}{l}{NLI} & Infer. & & 11,232 & 1,395 & 1,422 & \begin{tabular}{@{}c@{}} 263,029 / 29,352 \\ 18 \end{tabular} & \begin{tabular}{@{}l@{}}Clinical \\ notes\end{tabular} \\
\cline{2-9}
& \multicolumn{1}{l}{NER} & NER & \begin{tabular}{c}Acc \\ F1\end{tabular} & 3,440 & 676 & 693 & \begin{tabular}{@{}c@{}} 68,041 / 4,809 \\ 14 \end{tabular} & \begin{tabular}{@{}l@{}}User \\ reviews\end{tabular} \\
\hline
\end{tabular}
\end{table}

\subsection{RuMedTop3} \label{top3_sect}
In a medical code assignment task~\cite{crammer2007automatic}, a set of codes is supposed to be assigned to a raw medical report for analytical, insurance, or billing purposes. In other words, the task concept is multiple answers, multiple predictions. In our task formulation, the concept is a single answer, multiple predictions. A sample implies the only ground truth disease code, but we expect the three most probable ICD-10 codes. Therefore based only on raw text from a patient, the model should produce a ranked list of 3 codes. The model that decently solves this task has multiple practical applications, for example, the second opinion component in a clinical decision support system or part of an automatic patient triage system.

Both \emph{RuMedTop3} and \emph{RuMedSymptomRec} tasks are based on \emph{RuMedPrime} data~\cite{starovoytova}. The dataset contains 7,625 anonymized visit records from an outpatient unit of Siberian State Medical University hospital. Each record is represented with several columns. We are interested in the text field \emph{symptoms} and \emph{ICD10}. We use only the second level of the ICD-10 classification code hierarchy to prepare the target and drop the records with rare codes (under threshold 10). Thus only 6,360 visits have left with 105 target codes. Formally the task is defined as multi-class classification. Table~\ref{task_examples} gives examples of input data and labeling for this and the following tasks. 

For the evaluation, we propose to use \emph{Hit@1} (the same as \emph{Accuracy}) and \emph{Hit@3} metrics~\cite{schutze2008introduction}. The general formula is:
\begin{equation} \label{eq:1}
Hit@k = \frac{1}{N} \sum_{i=1}^{N} hit( \hat{y}_i, top_{i}^k ),
\end{equation}
\noindent where $N$ is the number of samples; $hit(\hat{y}, top^k)$ is 1 if ground truth ICD code $\hat{y}$ is on a ranked list of $k$ predicted codes $top^k$ and 0 otherwise.

The \emph{Hit@3} metric allows lessening the evaluation criterion as a decision about the diagnosis code has to be made in case of incomplete and limited information.

\subsection{RuMedSymptomRec} \label{rec_sect}
This task is designed to check models' ability to recommend a relevant symptom based on a given text premise. The task is beneficial in symptom checker applications where user interactions start with incomplete medical facts. Then, additional refining questions about possible user symptoms enable a more accurate diagnosis.

\begin{table}
\centering
\caption{Task examples from \emph{Dev} sets (in English for readability purpose\protect\footnotemark). Fragments in {\bf bold} are the labels, \email{monospaced} text represents field names, unformatted text is an input.} \label{task_examples}
\begin{tabular}{l m{0.17\linewidth} m{0.77\linewidth}}
\hline
\multirow{16}{*}{\rotatebox[origin=c]{90}{RuMed*}} & Top3 & \email{symptoms:} Palpitations, sleep disturbances, feeling short of breath. Pain and crunch in the neck, headaches for 3 days in a row.

\email{code:} {\bf M54} \\ 
\cline{2-3}
& SymptomRec & \email{symptoms:} The patient at the reception with relatives. According to relatives - complaints of poor sleep, a feeling of fear, obsessive thoughts that 'someone is beating her'

\email{code:} {\bf fluctuations in blood pressure} \\ 
\cline{2-3}
& DaNet & \email{context:} Epilepsy is a chronic polyetiological disease of the brain, the dominant manifestation of which is recurrent epileptic seizures resulting from an increased hypersynchronous discharge of brain neurons.

\email{question:} Is epilepsy a disease of the human brain?

\email{answer:} {\bf yes} \\ 
\cline{2-3}
& NLI & \email{ru\_sentence1:} During hospitalization, patient became progressively more dyspnic requiring BiPAP and then a NRB.

\email{ru\_sentence2:} The patient is on room air.

\email{gold\_label:} {\bf contradiction} \\ 
\cline{2-3}
& NER & $\stackrel[\email{ner\_tags:}]{}{\mathrm{\email{tokens:}}}$ $\stackrel[{\bf B-Drugname}]{}{\mathrm{Viferon}}$ $\stackrel[O]{}{\mathrm{has}}$ $\stackrel[O]{}{\mathrm{an}}$ $\stackrel[{\bf B-Drugclass}]{}{\mathrm{antiviral}}$ $\stackrel[O]{}{\mathrm{effect}}$ $\stackrel[O]{}{\mathrm{.}}$
\\ 
\hline
\end{tabular}
\end{table}
\footnotetext{For the original examples in Russian, please look at\\ \url{https://github.com/pavel-blinov/RuMedBench/blob/main/data/README.md}}

We define a \emph{symptom} as an explicit substring indicating a symptom-concept in the Unified Medical Language System (UMLS) metathesaurus~\cite{bodenreider2004unified}. Preliminary, we use the same \emph{symptoms}-field and our internal tool for symptoms extraction. Further, we select a random symptom as the target for each record and strip it from the initial text. After that, the task again transpires to multi-class classification with 141 symptom-codes. Finally, the same as \emph{RuMedTop3} metrics (\ref{eq:1}) are exploited for evaluation.

\subsection{RuMedDaNet} \label{danet_sect}
Our firm belief is that a genuinely medical AI model should have knowledge and "understanding" of different health-related domains. Partially this skill can be verified by checking the model's ability to answer context-specific yes/no questions. We designed the current task inspired by \emph{BoolQ}~\cite{clark2019boolq} and \emph{DaNetQA} from~\cite{russiansuperglue}.

A task sample consists of \emph{(context, question, answer)} triple. The context is an excerpt (up to 300 words long) from a medical-related text. We tried to collect the contexts from diverse fields: therapeutic medicine, human physiology and anatomy, pharmacology, biochemistry. In similar tasks, the questions are gathered from aggregated search engine queries. One disadvantage of such an approach is an explicit template-like question structure which produces unwanted lexical effects (like repeated patterns) and may shift the final assessment. We tried to avoid this issue and involved assessors for the questions generation from scratch.

First, the context passages are mined from open sources. Then given a context, an assessor was asked to generate a clarifying \emph{question} that can be definitely answered with either \emph{yes} or \emph{no}, thus completing triplet creation. All contexts are unique and paired with only one question. During the generation process, the balance between positive and negative questions is kept so that the \emph{Accuracy} metric is well suited for this task results' evaluation.

\subsection{RuMedNLI} \label{nli_sect}
The objective of the original NLI task is to compare a \emph{premise} and \emph{hypothesis} text and inference their relation as either \emph{entailment}, \emph{contradiction}, or \emph{neutral}. The paper~\cite{romanov2018lessons} proposed the medical formulation of the task called \emph{MedNLI}. The premises are extracted from the \emph{Past Medical History} section of MIMIC-III records; hypotheses are generated and labeled by clinicians. In the absence of similar data and labeling, we translated \emph{MedNLI} to Russian.

First, each text is independently processed by two automatic translation services. However, purely automatic results are poor because of domain-specific language and terminology; therefore, each sample needs a thorough second look by a human corrector to compile the final translation. Examples of such corrections include abbreviations and drug names adaptation, measurements conversion to the metric system units (Fahrenheit to Celsius, feet to meters, blood groups from ABO system to numeric, etc.), cultural and language phenomena replacement. The final counterpart \emph{RuMedNLI} dataset~\cite{rumednli22data} is available through the MIMIC-III derived data repository. Along with data, we keep the original data split and evaluation metric -- \emph{Accuracy}.

\subsection{RuMedNER} \label{ner_sect}
As mentioned in Section~\ref{relat_sect}, we included the NER task from~\cite{10.1093/bioinformatics/btaa675} in our benchmark. The data are user reviews about drug products. Each review is split into sentences and annotated with six types of named entities: drug name, class and form, adverse drug reactions, drug indications and symptoms of a disease, and "finding" (a miscellaneous type). The dataset with 4,566 entities is stored in IOB format (see Table~\ref{task_examples} for the example). Formally this is a multi-class per token classification task with \emph{Accuracy} and macro \emph{F1-score}~\cite{schutze2008introduction} evaluation metrics.

\section{Experiments}
\subsection{Baselines}
The most naive baseline method does not require modeling and is based solely on label statistics memorization. We estimate distributions in a \emph{Train} and select the most common labels as answers accordingly.

As more advanced baseline NLP methods, we selected the following ones.

{\bf Feature-based methods.} Statistical models are still decent first-choice options, especially with limited training data. In the case of the \emph{RuMedNER} task, we apply the Conditional Random Fields (CRF) model~\cite{lafferty2001conditional} with token-level features. In all other tasks for feature extraction, we use \emph{tf-idf} weighting scheme~\cite{schutze2008introduction} with analysis of char \emph{N}-grams ($N=3..8$). Then apply the logistic regression model~\cite{hastie_09_elements-of.statistical-learning} with the one-versus-all strategy for the final decision.

{\bf Recurrent Neural Networks (RNNs).} From the family of RNNs, we pick the Bidirectional Long Short-Term Memory (BiLSTM) architecture~\cite{963769}. In all tasks, we use a two-layer model with 300-dimensional word embeddings. We tried pre-trained word vectors as starting weights but, during experiments, found that random initialization yields better results.

{\bf Bidirectional Encoder Representations from Transformers (BERT).} 
We selected general domain RuBERT (12 layers, 12 self-attention heads, and 768 hidden layer size)~\cite{kuratov2019adaptation} as the base model for evaluation with the transformer models and RuPoolBERT~\cite{10.1007/978-3-030-59137-3_11} as its extended version. Both models fine-tuned for each task with the fixed set of hyperparameters, e.g., an input \(sequence~length\) is 256 tokens, 25 training \(epochs\) with a \(learning~rate\) of \num{3e-5}. The only task-specific parameter is the \(batch~size\). After the training, the best checkpoint was selected (regarding \emph{Dev} metrics) for the final prediction on the \emph{Test} set.

{\bf Human baseline.} To get an idea about human performance, we ask assessors to solve the tasks. A non-medical assessor solves the \emph{RuMedDaNet} and \emph{RuMedNER} task as they imply more common health-related knowledge and answers can be inferred from a context. Independent clinicians addressed the rest three tasks.

\subsection{Results and Discussion}
Table~\ref{bl_results} shows the performance results of tested baselines.

The naive method gains some performance in the tasks with unbalanced labels, e.g., \(Acc=10.58\) in the \emph{RuMedTop3} task means that more than 10\% of test samples belong to a most common ICD code. Nevertheless, the method is like a random guess for tasks with balanced label distribution (\emph{RuMedDaNet}, \emph{RuMedNLI}).

Generally, the transformer models outperform recurrent and feature-based ones. However, it is interesting to note the strong result of the linear model in the \emph{RuMedTop3} task. We attribute this to the simplicity of the task, e.g., from the statistical point of view, mapping between char \emph{N}-grams and a set of target codes is quite trivial. The method advantage diminishes with task complexity growth. For example, in the \emph{RuMedDaNet}, simple methods are close to naive ones. In contrast, BERT-like models fare above them because the task involves more knowledge and a deeper "understanding" of nuances between context and question. The BERT model with an advanced pooling strategy works better in classification and QA tasks but slightly worse in NER and NLI.

\begin{table}
\centering
\caption{Baseline performance metrics (\%) on the \emph{RuMedBench} test sets. The best models' results for each task are shown in {\bf bold}.} \label{bl_results}
\begin{tabular}{r c c c c c c}
& \multicolumn{5}{c}{RuMed*} & \\
\cline{2-6}
\multicolumn{1}{r}{Model} & $\stackrel{Acc/Hit@3}{\mathrm{Top3}}$ & $\stackrel{Acc/Hit@3}{\mathrm{SymptomRec}}$ & $\stackrel{Acc\textcolor{white}{/}}{\mathrm{DaNet}}$ & $\stackrel{Acc\textcolor{white}{/}}{\mathrm{NLI}}$ & $\stackrel{Acc/F1}{\mathrm{NER}}$ & Overall \\
\hline
\multicolumn{1}{r}{Naive} & 10.58/22.02 & 1.93/5.30 & 50.00 & 33.33 & 93.66/51.96 & 35.21 \\
\multicolumn{1}{r}{Feature-based} & {\bf 49.76}/{\bf 72.75} & 32.05/49.40 & 51.95 & 59.70 & 94.40/62.89 & 58.46 \\
\multicolumn{1}{r}{BiLSTM} & 40.88/63.50 & 20.24/31.33 & 52.34 & 60.06 & 94.74/63.26 & 53.87 \\
\multicolumn{1}{r}{RuBERT} & 39.54/62.29 & 18.55/34.22 & 67.19 & {\bf 77.64} & {\bf 96.63}/{\bf 73.53} & 61.44 \\
\multicolumn{1}{r}{RuPoolBERT} & 47.45/70.44 & {\bf 34.94}/{\bf 52.05} & {\bf 71.48} & 77.29 & ~96.47/73.15 & {\bf 67.20} \\
\hline
\multicolumn{1}{r}{Human} & 25.06/48.54 & 7.23/12.53 & 93.36 & 83.26 & 96.09/76.18 & 61.89 \\
\hline
\end{tabular}
\end{table}

Further, it is worth noting that models performances in the first two tasks are much better than clinicians. To clarify this finding, we discussed the result with the assessor-clinicians. During labeling, their main concern was insufficient and noisy information provided in the \emph{Symptom} field. Thus gender, age, physical examination results, and patient anamnesis are essential in medical diagnostic tasks; their exclusion leads to a substantial downgrade in human performance. Without such details, a model is a better conditional estimator of statistical distribution over many target classes.

Finally, human assessors hold the lead with substantial margins in the last three tasks. The \emph{RuMedDaNet} is the most challenging one of the proposed tasks, with the gap between the human level and the best model of more than 20\%. We hope the lag will be reduced after more advanced and specialized Russian medical models appear. Regarding the \emph{MedNLI} dataset, to the best of our knowledge, this is the first human performance assessment on the task, and the current model state-of-the-art result on its English edition is 86.57\%.

\subsection{Clinical Relevance}
We believe our experience outlined in this paper will provide a good reference for the research community developing AI methods in medicine. Specifically, the paper is notable either as an academic user case of non-English medical benchmark creation or starting point for Russian practitioners looking for open datasets and novel tasks. Overall, the proposed benchmark allows one to test Russian language models in the medical context and make better decisions before deployment.

\section{Conclusions}
In the study, we introduce the comprehensive open Russian medical language understanding benchmark, combining classification, QA, NLI, and NER tasks, four of them presented for the first time on newly created datasets. For comparison, we evaluate essential types of NLP methods and single-human performance. As expected, BERT-like models show the most promising results. However, even linear models can perform well for restricted, straightforward tasks. Moreover, in the current setup of medical diagnostic tasks, the models perform even better than humans. Therefore, future reformulations of such tasks should encompass more patient-related information, not only symptoms.

We hope our work will provide a sound basis for data practitioners and spur the research of text medical tasks in Russian. We release the source code and data in a unified format to start with the proposed benchmark quickly. Our further plan is the extension of \emph{RuMedBench} with more advanced tasks and the creation of a full-fledged evaluation platform with private tests.




\end{document}